\documentclass[11pt,letterpaper]{article}
\usepackage[auth-lg]{authblk}
\usepackage{emnlp2017}
\usepackage{times}
\usepackage{latexsym}

\emnlpfinalcopy

\def\emnlppaperid{***}

\usepackage{amsmath}
\usepackage{tabularx}
\usepackage{booktabs}
\usepackage{amssymb}
\usepackage{pifont}
\usepackage{bm}
\usepackage{multirow}
\usepackage{tcolorbox}
\usepackage[framemethod=tikz]{mdframed}
\usepackage{tikz,pgfplots}
\usepgfplotslibrary{groupplots}
\pgfplotsset{compat=1.13}

\newcommand\muc[0] {MUC}
\newcommand\bcubed[0] {$\text{B}^3$}
\newcommand\ceaf[0] {$\text{CEAF}_{\phi_4}$}
\newcommand\nl[1]{\textit{#1}}

\newcommand{\M}[2]{\mathbf{#1}_{\text{#2}}}
\newcommand{\V}[1]{\bm{#1}}
\newcommand{\mscore}[1]{s_\text{m}(#1)}
\newcommand{\ascore}[2]{s_\text{a}(#1, #2)}
\newcommand{\cscore}[2]{s(#1, #2)}
\newcommand{\ffnn}[2]{\textsc{ffnn}_\text{#1}(#2)}

\definecolor{g-red}{HTML}{DB4437}
\definecolor{g-blue}{HTML}{4285F4}
\definecolor{g-green}{HTML}{0F9D58}
\definecolor{g-yellow}{HTML}{F4B400}
\definecolor{g-orange}{HTML}{FF9800}
\definecolor{g-grey}{HTML}{9E9E9E}
\definecolor{uw}{RGB}{138,43,226}
\definecolor{stanford}{RGB}{255,69,0}
\definecolor{const}{RGB}{68, 110, 182}
\definecolor{head}{RGB}{246, 180, 32}
\definecolor{freq}{RGB}{0, 0, 0}

\newmdenv[innerlinewidth=0.5pt, roundcorner=4pt,linecolor=black,innerleftmargin=6pt,
innerrightmargin=6pt,innertopmargin=6pt,innerbottommargin=6pt]{examplebox}

\newcommand\cospan[1]{\textcolor{black}{\textbf{#1}}}
\newcommand\att[2]{\setlength{\fboxsep}{0.2pt}\colorbox{g-red!#1}{#2}}

\newcommand\layerbox[4]{
\draw[rounded corners] (#2, #3) rectangle (#2 + #1 * #4, #3 + #1 * 1);
}

\newcommand\layercolorbox[5][0.4] {
\draw[rounded corners, fill=#5] (#2, #3) rectangle (#2 + #1 * #4, #3 + #1 * 1);
}

\newcommand\layercomponent[5]{
\filldraw[fill=#5] (#2 + #1 * #4 - #1 * 0.5, #3 + #1 * 0.5) circle (#1 * 0.4);
}
\newcommand\layer[5][0.4] {
\layerbox{#1}{#2}{#3}{#4}
\foreach \x in {1, ..., #4}{
  \layercomponent{#1}{#2}{#3}{\x}{#5}
}
}

\newcommand\sumnode[3] {
\layercolorbox{#1-0.2}{#2-0.2}{1}{#3}
\node at (#1, #2) {\textcolor{white}{+}};
}

\newcommand\archcomment[3] {
\node[anchor=west, align=left] at (-3.7, #1) {\small\textbf{#2}};
}

\newcommand\archcommenttwo[4] {
\archcomment{#1}{#2}{#4}
\archcomment{#1-0.3}{#3}{#4}
}

\title{End-to-end Neural Coreference Resolution}

\author[$\dagger$]{Kenton Lee}
\author[$\dagger$]{Luheng He}
\author[$\ddagger$]{Mike Lewis}
\author[$\dagger$$*$]{Luke Zettlemoyer}
\affil[$\dagger$]{
Paul G. Allen School of Computer Science \& Engineering, Univ. of Washington, Seattle, WA  \authorcr
$^*$\hspace{4pt}Allen Institute for Artificial Intelligence, Seattle WA \authorcr
{\tt \{kentonl, luheng, lsz\}@cs.washington.edu}
\vspace*{6pt}}
\affil[$\ddagger$]{
Facebook AI Research\\
Menlo Park, CA \authorcr
{\tt mikelewis@fb.com}

}

\date{}

\begin{document}

\maketitle

\begin{abstract}
We introduce the first end-to-end coreference resolution model and show that it significantly outperforms all previous work without using a syntactic parser or hand-engineered mention detector. The key idea is to directly consider all spans in a document as potential mentions and learn distributions over possible antecedents for each. The model computes span embeddings that combine context-dependent boundary representations with a head-finding attention mechanism. It is trained to maximize the marginal likelihood of gold antecedent spans from coreference clusters and is factored to enable aggressive pruning of potential mentions. Experiments demonstrate state-of-the-art performance, with a gain of 1.5 F1 on the OntoNotes benchmark and by 3.1 F1 using a 5-model ensemble, despite the fact that this is the first approach to be successfully trained with no external resources.
\end{abstract}


\section{Introduction}

We present the first state-of-the-art neural coreference resolution model that is learned end-to-end given only gold mention clusters. All recent coreference models, including neural approaches that achieved impressive performance gains~\cite{wiseman:2016, clark:2016a, clark:2016b}, rely on syntactic parsers, both for head-word features and as the input to carefully hand-engineered mention proposal algorithms.
We demonstrate for the first time that these resources are not required, and in fact performance can be improved significantly without them, by training an end-to-end neural model that jointly learns which spans are entity mentions and how to best cluster them.

Our model reasons over the space of all spans up to a maximum length and directly optimizes the marginal likelihood of antecedent spans from gold coreference clusters. It includes a \emph{span}-ranking model that decides, for each span, which of the previous spans (if any) is a good antecedent.
At the core of our model are vector embeddings representing spans of text in the document, which combine context-dependent boundary representations with a head-finding attention mechanism over the span. The attention component is inspired by parser-derived head-word matching features from previous systems~\cite{durrett:2013}, but is less susceptible to cascading errors. In our analyses, we show empirically that these learned attention weights correlate strongly with traditional headedness definitions.

Scoring all span pairs in our end-to-end model is impractical, since the complexity would be quartic in the document length. 
Therefore we factor the model over unary mention scores and pairwise antecedent scores, both of which are simple functions of the learned span embedding. The unary mention scores are used to prune the space of spans and antecedents, to aggressively reduce the number of pairwise computations.

Our final approach outperforms existing models by 1.5 F1  on the OntoNotes benchmark and by 3.1 F1 using a 5-model ensemble. 
It is not only accurate, but also relatively interpretable.
The model factors, for example, directly indicate whether an absent coreference link is due to low mention scores (for either span) or a low score from the mention ranking component. 
The head-finding attention mechanism also reveals which mention-internal words contribute most to coreference decisions. We leverage this overall interpretability to do detailed quantitative and qualitative analyses, providing insights into the strengths and weaknesses of the approach.

\section{Related Work}
Machine learning methods have a long history in coreference resolution (see \newcite{ng:2010} for a detailed survey). However, the learning problem is challenging and, until very recently, hand-engineered systems built on top of automatically produced parse trees~\cite{raghunathan:2010} outperformed all learning approaches. \newcite{durrett:2013} showed that highly lexical learning approaches reverse this trend, and more recent neural models~\cite{wiseman:2016, clark:2016a, clark:2016b} have achieved significant performance gains. However, all of these models use parsers for head features and include highly engineered mention proposal algorithms.\footnote{For example,~\newcite{raghunathan:2010} use rules to remove pleonastic mentions of \nl{it} detected by 12 lexicalized regular expressions over English parse trees.} 
Such pipelined systems suffer from two major drawbacks: (1) parsing mistakes can introduce cascading errors and (2) many of the hand-engineered rules do not generalize to new languages.

A non-pipelined system that jointly models mention detection and coreference resolution was first proposed by~\newcite{daume2005large}. They introduce a search-based system that predicts the coreference structure in a left-to-right transition system that can incorporate global features. In contrast, our approach performs well while making much stronger independence assumptions, enabling straightforward inference.

More generally, a wide variety of approaches for learning coreference models have been proposed. They can typically be categorized as (1) mention-pair classifiers~\cite{ng:2002,bengtson:2008}, (2) entity-level models~\cite{haghighi:2010,clark:2015, clark:2016a, wiseman:2016}, (3) latent-tree models~\cite{fernandes:2012, bjorkelund:2014, martschat:2015}, or (4) mention-ranking models~\cite{durrett:2013,wiseman:2015,clark:2016b}. Our span-ranking approach is most similar to mention ranking, but we reason over a larger space by jointly detecting mentions and predicting coreference.

\newcommand\lstmnode[2] {
\filldraw[fill=g-yellow] (#1+0.2,#2+0.3) rectangle (#1+0.6,#2+0.7); 
}

\newcommand\forwardlstmconnect[3] {
\draw[-latex, line width=1pt] (#1+0.6,#2+0.25+0.7) -- (#1-0.3+#3,#2+0.25+0.7);
}
\newcommand\backwardlstmconnect[3] {
\draw[-latex, line width=1pt] (#1-0.3+#3,#2+0.25+0.7) -- (#1+0.6,#2+0.25+0.7);
}
\newcommand\lstmconnect[3] {
\forwardlstmconnect{#1}{#2+0.35}{#3}
\backwardlstmconnect{#1}{#2 + 1.15}{#3}
}

\newcommand\mentionout[4] {
\node[anchor=north, align=center] at (#1+0.6, 6.2) {$\text{\small #4}$}; 
\layer{#1}{4.6}{3}{g-green} 
\layer{#1+0.4}{5.3}{1}{black} 
\draw[-latex, line width=1pt] (#1+0.6, 5) to (#1+0.6, 5.3); 

\draw[-latex, line width=1pt, out=90, in=-90] (1.5 * #2 +0.4, 3.1) to (#1+0.2, 4.6); 

\draw[-latex, line width=1pt, out=90, in=-90] (1.5 * #3 + 0.4, 3.1) to (#1+1, 4.6); 

\sumnode{#1 + 0.6}{4.1}{g-red}

\foreach \x in {#2, ..., #3}{
\draw[-latex, line width=1pt, in=-90, out=90, looseness=1] (1.5 * \x + 0.4, 3.1) to (#1+0.6, 3.9); 
}
\draw[-latex, line width=1pt, in=-90, out=90] (#1+0.6, 4.3) to (#1+0.6, 4.6); 
}

\newcommand\word[2] {
\node[anchor=mid] at (#1+0.4, 0) {\small #2}; 
\layer{#1}{0.3}{2}{g-blue} 

\lstmnode{#1}{0.8}
\lstmnode{#1}{1.6}
\draw[-latex, line width=1pt] (#1+0.4, 0.7) to (#1+0.4, 1.1); 
\draw[-latex, line width=1pt, out=150, in=-150, looseness=1.5] (#1+0.4,0.7) to (#1+0.4, 1.9); 

\draw[-latex, line width=1pt, out=30, in=-30, looseness=1.5] (#1+0.4, 2 - 0.5) to (#1+0.4, 2 + 0.7); 

\draw[-latex, line width=1pt] (#1+0.4, 2.3) to (#1+0.4, 2.7); 

\layer{#1}{2.7}{2}{g-yellow} 
}

\begin{figure*}[ht!]
\begin{centering}
\scalebox{0.8} {
\begin{tikzpicture}
\word{0}{General}
\lstmconnect{0}{0}{2}
\word{1.5}{Electric}
\lstmconnect{1.5}{0}{2}
\word{3}{said}
\lstmconnect{3}{0}{2}
\word{4.5}{the}
\lstmconnect{4.5}{0}{2}
\word{6}{Postal}
\lstmconnect{6}{0}{2}
\word{7.5}{Service}
\lstmconnect{7.5}{0}{2}
\word{9}{contacted}
\lstmconnect{9}{0}{2}
\word{10.5}{the}
\lstmconnect{10.5}{0}{2}
\word{12}{company}

\mentionout{0.5}{0}{1}{General Electric}
\mentionout{2.8}{1}{3}{Electric said the}
\mentionout{5.8}{3}{5}{the Postal Service}
\mentionout{8.8}{5}{7}{Service contacted the}
\mentionout{11.2}{7}{8}{the company}

\archcomment{5.5}{Mention score ($s_\text{m}$)}{black}
\archcomment{4.8}{Span representation ($\V{g}$)}{g-green}
\archcomment{4.1}{Span head ($\hat{\V{x}}$)}{g-red}
\archcomment{2.9}{Bidirectional LSTM ($\V{x}^*$)}{g-yellow}
\archcommenttwo{0.6}{Word \& character}{embedding ($\V{x}$)}{g-blue}
\end{tikzpicture}
}
\caption{First step of the end-to-end coreference resolution model, which computes embedding representations of spans for scoring potential entity mentions. Low-scoring spans are pruned, so that only a manageable number of spans is considered for coreference decisions. In general, the model considers all possible spans up to a maximum width, but we depict here only a small subset. }
\label{fig:antecedent_viz}
\end{centering}
\end{figure*}
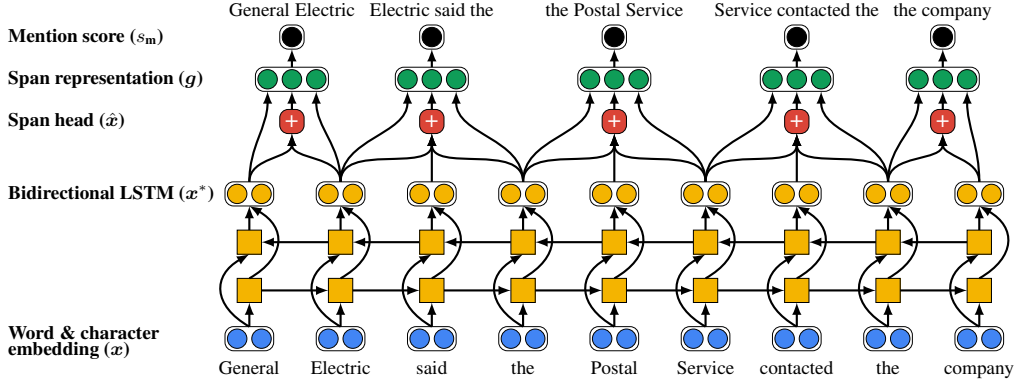

\newcommand\score[7] {
\node[anchor=#6, align=#7] at (#3, 7.9) {\small #4}; 
\node[anchor=#6, align=#7] at (#3, 7.6) {\small #5}; 
\layer{#3}{7.3}{1}{g-grey} 
\draw[-latex, line width=1pt] (#1+0.6, 5.7) to (#3+0.2, 7.3); 
\draw[-latex, line width=1pt] (#2+0.6, 5.7) to (#3+0.2, 7.3); 
\draw[-latex, line width=1pt] (#3+0.2, 6.4) to (#3+0.2, 7.3); 

\layer{#3}{6}{1}{white} 

\draw[-latex, line width=1pt] (#1+0.6, 4.9) to (#3+0.2, 6); 
\draw[-latex, line width=1pt] (#2+0.6, 4.9) to (#3+0.2, 6); 
}

\newcommand\mentionin[4] {
\node[anchor=north, align=center] at (#1+0.7, 4.4) {\small #4}; 

\layer{#1}{4.5}{3}{g-green} 
\layer{#1+0.4}{5.3}{1}{black} 
\draw[-latex, line width=1pt] (#1+0.6, 4.9) to (#1+0.6, 5.3); 
}

\begin{figure}[ht!]
\centering
\scalebox{0.8} {
\begin{tikzpicture}
\mentionin{-1}{0}{1}{General Electric}
\mentionin{1.3}{3}{5}{the Postal Service}
\mentionin{3.5}{7}{8}{the company}

\score{-1}{3.5}{1.2}{$s(\text{the company},\;\;\;\;\;\;$}{$\text{General Electric})$}{east}{right}
\score{1.3}{3.5}{2.3}{$\;\;\;\;s(\text{the company},$}{$\;\;\;\;\;\;\;\;\text{the Postal Service})$}{west}{left}
\node[anchor=north, align=center] at (0, 8.8) {\small$s(\text{the company},\epsilon)=0$}; 
\layer{0}{8.8}{1}{g-grey} 

\layer{1.3}{9}{3}{g-orange} 
\draw[-latex, line width=1pt] (0.4, 9) to (1.3,9.2);
\draw[-latex, line width=1pt] (1.4, 7.7) to (1.9,9);
\draw[-latex, line width=1pt] (2.5, 7.7) to (2.3,9);

\archcomment{9.1}{Softmax ($P(y_i \mid D)$)}{g-orange}
\archcommenttwo{7.6}{Coreference}{score ($s$)}{g-grey}
\archcomment{6.2}{Antecedent score ($s_\text{a}$)}{black}
\archcomment{5.5}{Mention score ($s_\text{m}$)}{black}
\archcommenttwo{4.8}{Span}{representation ($\V{g}$)}{g-green}
\end{tikzpicture}
}
\caption{Second step of our model. Antecedent scores are computed from pairs of span representations. The final coreference score of a pair of spans is computed by summing the mention scores of both spans and their pairwise antecedent score.
}
\label{fig:mention_viz}
\end{figure}
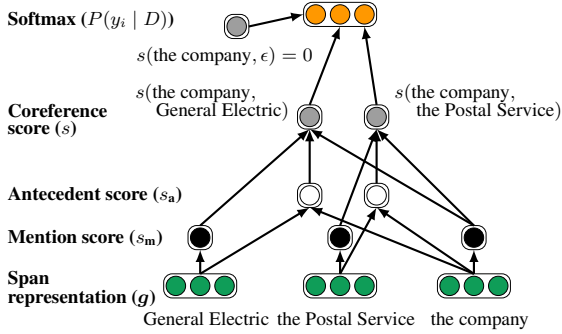

\section{Task}
We formulate the task of end-to-end coreference resolution as a set of decisions for every possible span in the document. The input is a document $D$ containing $T$ words along with metadata such as speaker and genre information.

Let $N = \frac{T (T + 1)}{2}$ be the number of possible text spans 
in $D$. Denote the start and end indices of a span $i$ in $D$ respectively by \textsc{start}(i) and \textsc{end}(i), for $1 \le i \le N$. We assume an ordering of the spans based on \textsc{start}(i); spans with the same start index are ordered by \textsc{end}(i).

The task is to assign to each span $i$ an antecedent $y_i$. The set of possible assignments for each $y_i$ is $\mathcal{Y}(i) = \{\epsilon, 1, \ldots, i - 1\}$, a dummy antecedent $\epsilon$ and all preceding spans.  True antecedents of span $i$, i.e. span $j$ such that $1 \le j \le i -1$, represent coreference links between $i$ and $j$. The dummy antecedent $\epsilon$ represents two possible scenarios: (1) the span is not an entity mention or (2) the span is an entity mention but it is not coreferent with any previous span.
These decisions implicitly define a final clustering, which can be recovered by grouping all spans that are connected by a set of antecedent predictions.  


\section{Model}
\label{sec:model}

We aim to learn a conditional probability distribution $P(y_1, \ldots, y_N \mid D)$ whose most likely configuration produces the correct clustering.
We use a product of multinomials for each span:
\begin{align*}
P(y_1, \ldots, y_N \mid D) &= \prod_{i=1}^N P(y_i \mid D)\\
&= \prod_{i=1}^N \frac{\exp(s(i, y_i))}{\sum_{y' \in \mathcal{Y}(i)}\exp(s(i, y'))}
\end{align*}
where $s(i, j)$ is a pairwise score for a coreference link between span $i$ and span $j$ in document $D$. We omit the document $D$ from the notation when the context is unambiguous. There are three factors for this pairwise coreference score: (1) whether span $i$ is a mention, (2) whether span $j$ is a mention, and (3) whether $j$ is an antecedent of $i$:
\begin{align*}
    \cscore{i}{j} &= \begin{cases}
    0 & j= \epsilon\\
    \mscore{i} + \mscore{j} + \ascore{i}{j} & j \neq \epsilon
    \end{cases}
\end{align*}
Here $\mscore{i}$ is a unary score for span $i$ being a mention, and $\ascore{i}{j}$ is pairwise score for span $j$ being an antecedent of span $i$. 

By fixing the score of the dummy antecedent $\epsilon$ to $0$, the model predicts the best scoring antecedent if any non-dummy scores are positive, and it abstains if they are all negative.

A challenging aspect of this model is that its size is $\mathcal{O}(T^4)$ in the document length. As we will see in Section~\ref{sec:inference}, the above factoring enables aggressive pruning of spans that are unlikely to belong to a coreference cluster according the mention score $\mscore{i}$.

\paragraph{Scoring Architecture}
We propose an end-to-end neural architecture that computes the above scores given the document and its metadata.

At the core of the model are vector representations $\V{g}_i$ for each possible span $i$, which we describe in detail in the following section. Given these span representations, the scoring functions above are computed via standard feed-forward neural networks:
\begin{align*}
    \mscore{i}&= \V{w}_\text{m} \cdot \ffnn{m}{\V{g}_i}\\
    \ascore{i}{j} &= \V{w}_\text{a} \cdot \ffnn{a}{[\V{g}_i, \V{g}_j, \V{g}_i \circ \V{g}_j, \phi(i, j)]}
\end{align*}
\noindent
where $\cdot$ denotes the dot product, $\circ$ denotes element-wise multiplication, and $\textsc{ffnn}$ denotes a feed-forward neural network that computes a non-linear mapping from input to output vectors.

The antecedent scoring function $\ascore{i}{j}$ includes explicit element-wise similarity of each span $\V{g}_i \circ \V{g}_j$ and a feature vector $\phi(i, j)$ encoding speaker and genre information from the metadata and the distance between the two spans.

\paragraph{Span Representations}
Two types of information are crucial to accurately predicting coreference links: the context surrounding the mention span and the internal structure within the span. We use a bidirectional LSTM~\cite{lstm} to encode the lexical information of both the inside and outside of each span. We also include an attention mechanism over words in each span to model head words.

We assume vector representations of each word $\{\V{x}_1, \ldots, \V{x}_T\}$, which are composed of fixed pre-trained word embeddings and 1-dimensional convolution neural networks (CNN) over characters (see Section~\ref{sec:hyperparameters} for details)

To compute vector representations of each span, we first use bidirectional LSTMs to encode every word in its context:
\begin{align*}
\V{f}_{t, \delta}&=\sigma(\M{W}{f} [\V{x}_t, \V{h}_{t+\delta, \delta}] + \V{b}_\text{i})\\
\V{o}_{t, \delta}&=\sigma(\M{W}{o} [\V{x}_t, \V{h}_{t+\delta, \delta}] + \V{b}_\text{o})\\
\V{\widetilde{c}}_{t, \delta} &= \tanh(\M{W}{c}[\V{x}_t, \V{h}_{t+\delta, \delta}]+\V{b}_\text{c})\\
\V{c}_{t, \delta}&=\V{f}_{t, \delta}\circ \V{\widetilde{c}}_{t, \delta}+(\V{1}-\V{f}_{t, \delta}) \circ \V{c}_{t + \delta, \delta}\\
\V{h}_{t, \delta}&=\V{o}_{t, \delta}\circ \tanh(\V{c}_{t, \delta})\\
\V{x}^*_t &= [\V{h}_{t, 1}, \V{h}_{t, -1}]
\end{align*}
where $\delta \in \{-1, 1\}$ indicates the directionality of each LSTM, and $\V{x}^*_t$ is the concatenated output of the bidirectional LSTM. We use independent LSTMs for every sentence, since cross-sentence context was not helpful in our experiments.

Syntactic heads are typically included as features in previous systems~\cite{durrett:2013, clark:2016a, clark:2016b}. Instead of relying on syntactic parses, our model learns a task-specific notion of headedness using an attention mechanism \cite{bahdanau:2014} over words in each span:
\begin{align*}
\alpha_t &= \V{w}_\alpha \cdot \ffnn{$\alpha$}{\V{x}^*_t}\\
a_{i,t} &= \frac{\exp(\alpha_t)}{\displaystyle\sum_{k=\textsc{start}(i)}^{\textsc{end}(i)} \exp(\alpha_k)}\\
\hat{\V{x}}_i &= \sum_{t=\textsc{start}(i)}^{\textsc{end}(i)} a_{i, t} \cdot  \V{x}_t
\end{align*}
where $\hat{\V{x}}_i$ is a weighted sum of word vectors in span $i$. The weights $a_{i,t}$ are automatically learned and correlate strongly with traditional definitions of head words as we will see in Section~\ref{sec:precision}.

The above span information is concatenated to produce the final representation $\V{g}_i$ of span $i$:
\begin{align*}
\V{g}_i &= [\V{x}^*_{\textsc{start}(i)}, \V{x}^*_{\textsc{end}(i)}, \hat{\V{x}}_i, \phi(i)]
\end{align*}
This generalizes the recurrent span representations recently proposed for question-answering~\cite{lee:2016}, which only include the boundary representations $\V{x}^*_{\textsc{start}(i)}$ and $\V{x}^*_{\textsc{end}(i)}$. We introduce the soft head word vector $\hat{\V{x}}_i$ and a feature vector $\phi(i)$ encoding the size of span $i$.

\section{Inference}
\label{sec:inference}
The size of the full model described above is $\mathcal{O}(T^4)$ in the document length $T$. To maintain computation efficiency, we prune the candidate spans greedily during both training and evaluation.

We only consider spans with up to $L$ words and compute their unary mention scores $\mscore{i}$ (as defined in Section~\ref{sec:model}). To further reduce the number of spans to consider, we only keep up to $\lambda T$ spans with the highest mention scores and consider only up to $K$ antecedents for each. 
We also enforce non-crossing bracketing structures with a simple suppression scheme.\footnote{The official CoNLL-2012 evaluation only considers predictions without crossing mentions to be valid. Enforcing this consistency is not inherently necessary in our model.}
We accept spans in decreasing order of the mention scores, unless, when considering span $i$, there exists a previously accepted span $j$ such that  $\textsc{start}(i) < \textsc{start}(j) \le  \textsc{end}(i) < \textsc{end}(j) \lor \textsc{start}(j) < \textsc{start}(i) \le \textsc{end}(j) < \textsc{end}(i)$. 


Despite these aggressive pruning strategies, we maintain a high recall of gold mentions in our experiments (over 92\% when $\lambda=0.4$).

For the remaining mentions, the joint distribution of antecedents for each document is computed in a forward pass over a single computation graph. The final prediction is the clustering produced by the most likely configuration.

\section{Learning}
\label{sec:learning}
In the training data, only clustering information is observed. Since the antecedents are latent, we optimize the marginal log-likelihood of all correct antecedents implied by the gold clustering:
\begin{align*}
\log \prod_{i=1}^N \sum_{\hat{y} \in \mathcal{Y}(i) \cap \textsc{gold}(i)}P(\hat{y})
\end{align*}
where $\textsc{gold}(i)$ is the set of spans in the gold cluster containing span $i$. If span $i$ does not belong to a gold cluster or all gold antecedents have been pruned, $\textsc{gold}(i) = \{\epsilon\}$. 

By optimizing this objective, the model naturally learns to prune spans accurately. While the initial pruning is completely random, only gold mentions receive positive updates. The model can quickly leverage this learning signal for appropriate credit assignment to the different factors, such as the mention scores $s_m$ used for pruning.

Fixing score of the dummy antecedent to zero removes a spurious degree of freedom in the overall model with respect to mention detection. It also prevents the span pruning from introducing noise. 
For example, consider the case where span $i$ has a single gold antecedent that was pruned, so $\textsc{gold}(i) = \{\epsilon\}$. The learning objective will only correctly push the scores of non-gold antecedents lower, and it cannot incorrectly push the score of the dummy antecedent higher.

This learning objective can be considered a span-level, cost-insensitive analog of the learning objective proposed by~\newcite{durrett:2013}. We experimented with these cost-sensitive alternatives, including margin-based variants~\cite{wiseman:2015, clark:2016b}, but a simple maximum-likelihood objective proved to be most effective.
\begin{table*}[ht!]
\setlength{\tabcolsep}{0.3em}
\centering
\begin{tabularx}{\linewidth}{X c*{14}{c}}
\toprule
& \multicolumn{4}{c}{MUC} &  \multicolumn{4}{c}{\bcubed} & \multicolumn{4}{c}{\ceaf} & \\ 
 & Prec.  & Rec.  & F1 & \; & Prec. & Rec. & F1 & \;  & Prec.  & Rec.  & F1 & \; &  \multicolumn{1}{c}{Avg. F1} \\
\midrule
Our model (ensemble)   & \textbf{81.2} & \textbf{73.6} & \textbf{77.2} && \textbf{72.3} & \textbf{61.7} & \textbf{66.6} && \textbf{65.2} & \textbf{60.2} & \textbf{62.6} && \textbf{68.8} \\
Our model (single)   & 78.4 & 73.4 & 75.8 && 68.6 & 61.8 & 65.0 && 62.7 & 59.0 & 60.8 && 67.2\\
\cmidrule{1-14}
\newcite{clark:2016b}     & 79.2 & 70.4 & 74.6 && 69.9 & 58.0 & 63.4 && 63.5 & 55.5 & 59.2 && 65.7\\ 
\newcite{clark:2016a}     & 79.9 & 69.3 & 74.2 && 71.0 & 56.5 & 63.0 && 63.8 & 54.3 & 58.7 && 65.3\\ 
\newcite{wiseman:2016}    & 77.5 & 69.8 & 73.4 && 66.8 & 57.0 & 61.5 && 62.1 & 53.9 & 57.7 && 64.2\\
\newcite{wiseman:2015}    & 76.2 & 69.3 & 72.6 && 66.2 & 55.8 & 60.5 && 59.4 & 54.9 & 57.1 && 63.4\\
\newcite{clark:2015}      & 76.1 & 69.4 & 72.6 && 65.6 & 56.0 & 60.4 && 59.4 & 53.0 & 56.0 && 63.0\\
\newcite{martschat:2015}  & 76.7 & 68.1 & 72.2 && 66.1 & 54.2 & 59.6 && 59.5 & 52.3 & 55.7 && 62.5\\
\newcite{durrett:2014}    & 72.6 & 69.9 & 71.2 && 61.2 & 56.4 & 58.7 && 56.2 & 54.2 & 55.2 && 61.7\\
\newcite{bjorkelund:2014} & 74.3 & 67.5 & 70.7 && 62.7 & 55.0 & 58.6 && 59.4 & 52.3 & 55.6 && 61.6\\
\newcite{durrett:2013}    & 72.9 & 65.9 & 69.2 && 63.6 & 52.5 & 57.5 && 54.3 & 54.4 & 54.3 && 60.3\\
\bottomrule
\end{tabularx}

\caption{Results on the test set on the English data from the CoNLL-2012 shared task. The final column (Avg. F1) is the main evaluation metric, computed by averaging the F1 of \muc, \bcubed, and \ceaf. We improve state-of-the-art performance by 1.5 F1 for the single model and by 3.1 F1.}
\label{tab:results}
\end{table*}
\section{Experiments}
We use the English coreference resolution data from the CoNLL-2012 shared task~\cite{pradhan:2012} in our experiments. This dataset contains 2802 training documents, 343 development documents, and 348 test documents. The training documents contain on average 454 words and a maximum of 4009 words.

\subsection{Hyperparameters}
\label{sec:hyperparameters}
\paragraph{Word representations}
The word embeddings are a fixed concatenation of 300-dimensional GloVe embeddings~\cite{pennington:2014} and 50-dimensional embeddings from~\newcite{turian:2010}, both normalized to be unit vectors. Out-of-vocabulary words are represented by a vector of zeros. In the character CNN, characters are represented as learned 8-dimensional embeddings. The convolutions have window sizes of 3, 4, and 5 characters, each consisting of 50 filters.

\paragraph{Hidden dimensions} The hidden states in the LSTMs have 200 dimensions. Each feed-forward neural network consists of two hidden layers with 150 dimensions and rectified linear units~\cite{relu}.

\paragraph{Feature encoding} We encode speaker information as a binary feature indicating whether a pair of spans are from the same speaker. Following \newcite{clark:2016a}, the distance features are binned into the following buckets [1, 2, 3, 4, 5-7, 8-15, 16-31, 32-63, 64+]. All features (speaker, genre, span distance, mention width) are represented as learned 20-dimensional embeddings.

\paragraph{Pruning} We prune the spans such that the maximum span width $L=10$, the number of spans per word $\lambda=0.4$, and the maximum number of antecedents $K=250$. During training, documents are randomly truncated to up to 50 sentences.

\paragraph{Learning} We use ADAM~\cite{kingma:2016} for learning with a minibatch size of 1. The LSTM weights are initialized with random orthonormal matrices as described in \newcite{saxe:2013}. We apply 0.5 dropout to the word embeddings and character CNN outputs. We apply 0.2 dropout to all hidden layers and feature embeddings. Dropout masks are shared across timesteps to preserve long-distance information as described in \newcite{gal:2016}. The learning rate is decayed by 0.1\% every 100 steps. The model is trained for up to 150 epochs, with early stopping based on the development set.

All code is implemented in TensorFlow~\cite{tensorflow} and is publicly available. \footnote{\url{https://github.com/kentonl/e2e-coref}}

\subsection{Ensembling}
We also report ensemble experiments using five models trained with different random initializations.
Ensembling is performed for both the span pruning and antecedent decisions.

At test time, we first average the mention scores $\mscore{i}$ over each model before pruning the spans. Given the same pruned spans, each model then computes the antecedent scores $\ascore{i}{j}$ separately, and they are averaged to produce the final scores.

\section{Results}
We report the precision, recall, and F1 for the standard \muc, \bcubed, and \ceaf metrics using the official CoNLL-2012 evaluation scripts. The main evaluation is the average F1 of the three metrics.

\subsection{Coreference Results}
Table~\ref{tab:results} compares our model to several previous systems that have driven substantial improvements over the past several years on the OntoNotes benchmark. We outperform previous systems in all metrics. In particular, our single model improves the state-of-the-art average F1 by 1.5, and our 5-model ensemble improves it by 3.1.

The most significant gains come from improvements in recall, which is likely due to our end-to-end setup. During training, pipelined systems typically discard any mentions that the mention detector misses, which for~\newcite{clark:2016b} consists of more than 9\% of the labeled mentions in the training data. In contrast, we only discard mentions that exceed our maximum mention width of 10, which accounts for less than 2\% of the training mentions. The contribution of joint mention scoring is further discussed in Section~\ref{sec:span_pruning}

\subsection{Ablations}
\begin{table}[t!]
\newcolumntype{Y}{>{\centering\arraybackslash}X}
\newcommand{\colindent}{\;}
\setlength{\tabcolsep}{0.25em}
\centering
\begin{tabularx}{\linewidth}{l Y Y}
\toprule
 & Avg.~F1 & $\Delta$ \\
\midrule
Our model (ensemble) &  69.0 & +1.3\\
Our model (single) & 67.7 &  \\
\colindent $-$ distance and width features & 63.9 & -3.8\\
\colindent $-$ GloVe embeddings & 65.3 & -2.4\\
\colindent $-$ speaker and genre metadata & 66.3 & -1.4\\
\colindent $-$ head-finding attention & 66.4 & -1.3\\
\colindent $-$ character CNN & 66.8 & -0.9 \\
\colindent $-$ Turian embeddings & 66.9 & -0.8 \\
\bottomrule
\end{tabularx}

\caption{Comparisons of our single model on the development data. The 5-model ensemble provides a 1.3 F1 improvement. The head-finding attention, features, and all word representations contribute significantly to the full model.}
\label{tab:ablations}
\end{table}
To show the importance of each component in our proposed model, we ablate various parts of the architecture and report the average F1 on the development set of the data (see Figure~\ref{tab:ablations}).

\paragraph{Features} The distance between spans and the width of spans are crucial signals for coreference resolution, consistent with previous findings from other coreference models. They contribute 3.8 F1 to the final result. 

\paragraph{Word representations} Since our word embeddings are fixed, having access to a variety of word embeddings allows for a more expressive model without overfitting. We hypothesis that the different learning objectives of the GloVe and Turian embeddings provide orthogonal information (the former is word-order insensitive while the latter is word-order sensitive). Both embeddings contribute to some improvement in development F1.

The character CNN provides morphological information and a way to backoff for out-of-vocabulary words. Since coreference decisions often involve rare named entities, we see a contribution of 0.9 F1 from character-level modeling.

\paragraph{Metadata} Speaker and genre indicators many not be available in downstream applications.  We show that performance degrades by 1.4 F1 without them, but is still on par with previous state-of-the-art systems that assume access to this metadata.

\paragraph{Head-finding attention} Ablations also show a 1.3 F1 degradation in performance without the attention mechanism for finding task-specific heads. As we will see in Section~\ref{sec:examples}, the attention mechanism should not be viewed as simply an approximation of syntactic heads. In many cases, it is beneficial to pay attention to multiple words that are useful specifically for coreference but are not traditionally considered to be syntactic heads.

\subsection{Comparing Span Pruning Strategies}
\label{sec:span_pruning}
\begin{table}[t!]
\newcolumntype{Y}{>{\centering\arraybackslash}X}
\newcommand{\colindent}{\;}
\setlength{\tabcolsep}{0.25em}
\centering
\begin{tabularx}{\linewidth}{l Y Y}
\toprule
 & Avg.~F1 & $\Delta$ \\
\midrule
Our model (joint mention scoring) & 67.7 &  \\
\colindent w/ rule-based mentions & 66.7 & -1.0\\
\colindent w/ oracle mentions & 85.2 & +17.5\\
\bottomrule
\end{tabularx}

\caption{Comparisons of of various mention proposal methods with our model on the development data. The rule-based mentions are derived from the mention detector from~\newcite{raghunathan:2010}, resulting in a 1 F1 drop in performance. The oracle mentions are from the labeled clusters and improve our model by over 17.5 F1.}
\label{tab:span_pruning}
\end{table}
To tease apart the contributions of improved mention scoring and improved coreference decisions, we compare the results of our model with alternate span pruning strategies. In these experiments, we use the alternate spans for both training and evaluation. As shown in Table~\ref{tab:span_pruning}, keeping mention candidates detected by the rule-based system over predicted parse trees~\cite{raghunathan:2010} degrades performance by 1 F1. We also provide oracle experiment results, where we keep exactly the mentions that are present in gold coreference clusters. With oracle mentions, we see an improvement of 17.5 F1, suggesting an enormous room for improvement if our model can produce better mention scores and anaphoricity decisions.

\section{Analysis}
To highlight the strengths and weaknesses of our model, we provide both quantitative and qualitative analyses. In the following discussion, we use predictions from the single model rather than the ensembled model.

\begin{figure}[t!]
\vspace{-7pt}
\begin{tikzpicture}
\begin{axis}[
    	width=1.0\columnwidth,
	    height=0.8\columnwidth,
	    legend cell align=left,
	    legend style={at={(1, 0)},anchor=south east,font=\small},
	    xtick={0.1, 0.2, 0.3, 0.4, 0.5},
   	 	ytick={10, 20, 30, 40, 50, 60, 70, 80, 90, 100},
   		ymin=50, ymax=100,
   		xtick pos=left,
   		xtick align=outside,
	    xmin=0.05,xmax=0.55,
	    mark options={mark size=2},
		font=\small,
   	 	ymajorgrids=true,
    	xmajorgrids=true,
    	xlabel=Spans per word $\lambda$,
        ylabel=Recall (\%),
    	ylabel style={yshift=-1ex,}]
\addplot[
    color=uw,
    line width=1.5pt
    ]
    coordinates {
(0.1, 54.81)(0.15, 73.72)(0.2, 83.39)(0.25, 87.99)(0.3,90.20)(0.4,92.67)(0.5, 93.78)
    };
    \addlegendentry{Our model (various $\lambda$)}
\addplot[
    only marks,
    color=uw,
    mark=*,
    mark size=10pt
    ]
    coordinates {
(0.4,92.67)
    };
    \addlegendentry{Our model (actual $\lambda$)}
\addplot[
    only marks,
    color=stanford,
    mark=square*,
    mark size=10pt
    ]
    coordinates {
    (0.264829126306, 89.24)
    };
    \addlegendentry{\newcite{raghunathan:2010}}
\end{axis}
\end{tikzpicture}
\caption{Proportion of gold mentions covered in the development data as we increase the number of spans kept per word. Recall is comparable to the mention detector of previous state-of-the-art systems given the same number of spans. Our model keeps 0.4 spans per word in our experiments, achieving 92.7\% recall of gold mentions.}
\label{fig:recall}
\end{figure}
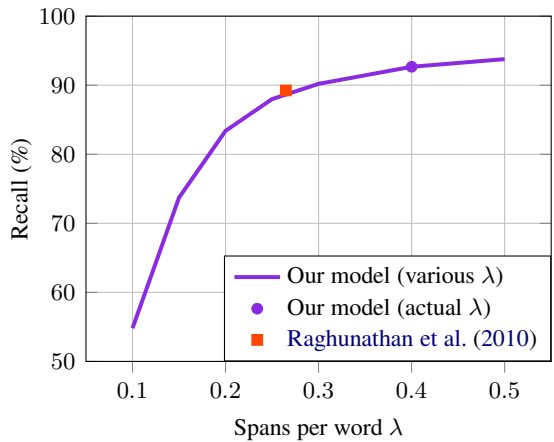
\subsection{Mention Recall}
\label{sec:recall}
The training data only provides a weak signal for spans that correspond to entity mentions, since singleton clusters are not explicitly labeled. As a by product of optimizing marginal likelihood, our model automatically learns a useful ranking of spans via the unary mention scores from Section~\ref{sec:model}.

The top spans, according to the mention scores, cover a large portion of the mentions in gold clusters, as shown in Figure~\ref{fig:recall}. Given a similar number of spans kept, our recall is comparable to the rule-based mention detector~\cite{raghunathan:2010} that produces 0.26 spans per word with a recall of 89.2\%. As we increase the number of spans per word ($\lambda$ in Section~\ref{sec:inference}), we observe higher recall but with diminishing returns. In our experiments, keeping 0.4 spans per word results in 92.7\% recall in the development data.

\subsection{Mention Precision}
\label{sec:precision}
While the training data does not offer a direct measure of mention precision, we can use the gold syntactic structures provided in the data as a proxy. Spans with high mention scores should correspond to syntactic constituents.

\begin{figure}[t!]
    \centering
        \begin{tikzpicture}[trim left=-0.6cm]
        \begin{axis}[
    	width=1.05\columnwidth,
	    height=0.8\columnwidth,
	    legend cell align=left,
	    legend style={at={(1, 1)},anchor=north east,font=\small},
        xtick={1, 2, 3, 4, 5, 6, 7, 8, 9, 10},
   	 	ytick={10, 20, 30, 40, 50, 60, 70, 80, 90, 100},
   		ymin=0, ymax=120,
	    xtick pos=left,
        ybar,
    	ymajorgrids=true,
		font=\small,
        bar width=0.2,
    	xlabel=Span width,
        ylabel=\%,
    	ylabel style={yshift=-1ex,},
        ]
    \addplot[fill=const] coordinates {
(1,100.0)
(2,89.8261898262)
(3,86.4031351167)
(4,75.9782608696)
(5,69.5683196575)
(6,61.9980879541)
(7,57.3922531369)
(8,52.0740740741)
(9,46.2601626016)
(10,45.6026058632)
            };
    \addlegendentry{Constituency precision}
            
    \addplot[fill=head] coordinates {
(1,100.0)
(2,92.8781278493)
(3,86.0974166831)
(4,78.0400572246)
(5,78.5641025641)
(6,78.0262143408)
(7,69.8669201521)
(8,71.9772403983)
(9,68.1898066784)
(10,68.0952380952)
            };
    \addlegendentry{Head word precision}
    
    \addplot[fill=freq] coordinates {
(1, 59.955)
(2, 20.589)
(3, 8.079)
(4, 4.053)
(5, 2.619)
(6, 1.391)
(7, 1.218)
(8, 0.905)
(9, 0.722)
(10, 0.469)
            };
    \addlegendentry{Frequency}
    
    \end{axis}
    \end{tikzpicture}
    \caption{Indirect measure of mention precision using agreement with gold syntax. Constituency precision: \% of unpruned spans matching syntactic constituents. Head word precision: \% of unpruned constituents whose syntactic head word matches the most attended word. Frequency: \% of gold spans with each width.}
    \label{fig:precision}
\end{figure}
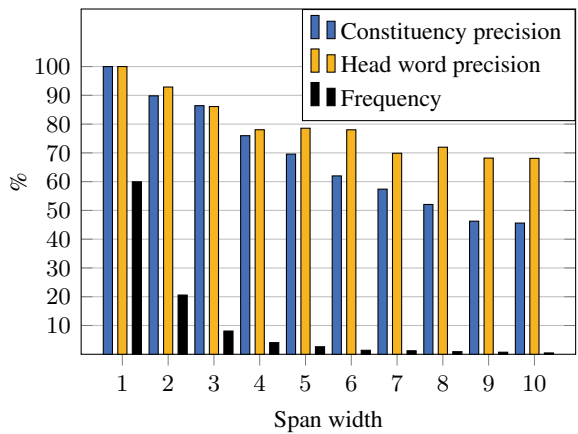
In Figure~\ref{fig:precision}, we show the precision of top-scoring spans when keeping 0.4 spans per word. For spans with 2--5 words, 75--90\% of the predictions are constituents, indicating that the vast majority of the mentions are syntactically plausible. Longer spans, which are all relatively rare, prove more difficult for the model, and precision drops to 46\% for spans with 10 words.

\begin{table*}[t!]
\setlength{\tabcolsep}{0.3em}
\centering
\fontsize{10.5}{12.6}\selectfont
\def\tabularxcolumn#1{m{#1}}
\begin{tabularx}{\linewidth}{cX}
\toprule
\multirow{3}{*}{\vspace{-5pt}1} & \cospan{(\att{3}{A} \att{90}{fire} \att{3}{in} \att{0}{a} \att{0}{Bangladeshi} \att{1}{garment} \att{2}{factory})} has left at least 37 people dead and 100 hospitalized. Most of the deceased were killed in the crush as workers tried to flee \cospan{(\att{0}{the} \att{100}{blaze})} in the four-story building. \\
\cmidrule{2-2}
 &A fire in \cospan{(\att{3}{a} \att{11}{Bangladeshi} \att{22}{garment} \att{64}{factory})} has left at least 37 people dead and 100 hospitalized. Most of the deceased were killed in the crush as workers tried to flee the blaze in \cospan{(\att{1}{the} \att{5}{four}\att{1}{-}\att{5}{story} \att{88}{building})}. \\
 \midrule
2 & We are looking for \cospan{(\att{1}{a} \att{72}{region} \att{1}{of} \att{8}{central} \att{4}{Italy} \att{1}{bordering} \att{0}{the} \att{2}{Adriatic} \att{3}{Sea})}. \cospan{(\att{0}{The} \att{100}{area})} is mostly mountainous and includes Mt. Corno, the highest peak of the Apennines. \cospan{(\att{100}{It})} also includes a lot of sheep, good clean-living, healthy sheep, and an Italian entrepreneur has an idea about how to make a little money of them.\\
\midrule
3 & \cospan{(\att{0}{The} \att{13}{flight} \att{87}{attendants})} have until 6:00 today to ratify labor concessions. \cospan{(\att{1}{The} \att{98}{pilots}\att{1}{'})} union and ground crew did so yesterday. \\
\midrule
4 & \cospan{(\att{2}{Prince} \att{3}{Charles} \att{77}{and} \att{0}{his} \att{0}{new} \att{3}{wife} \att{14}{Camilla})} have jumped across the pond and are touring the United States making \cospan{(\att{100}{their})} first stop today in New York. It's Charles' first opportunity to showcase his new wife, but few Americans seem to care. Here's Jeanie Mowth. What a difference two decades make. \cospan{(\att{15}{Charles} \att{64}{and} \att{22}{Diana})} visited a JC Penney's on the prince's last official US tour. Twenty years later here's the prince with his new wife. \\
\midrule
5 & Also such location devices, \cospan{(\att{29}{some} \att{71}{ships})} have smoke floats \cospan{(\att{100}{they})} can toss out so the man overboard will be able to use smoke signals as a way of trying to, let the rescuer locate \cospan{(\att{100}{them})}.\\
\bottomrule
\end{tabularx}
\caption{Examples predictions from the development data. Each row depicts a single coreference cluster predicted by our model. Bold, parenthesized spans indicate mentions in the predicted cluster. The redness of each word indicates the weight of the head-finding attention mechanism ($a_{i,t}$ in Section~\ref{sec:model}).}
\label{tab:examples}
\end{table*}
\subsection{Head Agreement}
We also investigate how well the learned head preferences correlate with syntactic heads. For each of the top-scoring spans in the development data that correspond to gold constituents, we compute the word with the highest attention weight.

We plot in Figure~\ref{fig:precision} the proportion of these words that match syntactic heads. Agreement ranges between 68-93\%, which is surprisingly high, since no explicit supervision of syntactic heads is provided. The model simply learns from the clustering data that these head words are useful for making coreference decisions.

\subsection{Qualitative Analysis}
\label{sec:examples}

Our qualitative analysis in Table~\ref{tab:examples} highlights the strengths and weaknesses of our model. Each row is a visualization of a single coreference cluster predicted by the model. Bolded spans in parentheses belong to the predicted cluster, and the redness of a word indicates its weight from the head-finding attention mechanism ($a_{i,t}$ in Section~\ref{sec:model}).

\paragraph{Strengths}
The effectiveness of the attention mechanism for making coreference decisions can be seen in Example 1. The model pays attention to \nl{fire} in the span \nl{A fire in a Bangladeshi garment factory}, allowing it to successfully predict the coreference link with \nl{the blaze}. For a subspan of that mention, \nl{a Bangladeshi garment factory}, the model pays most attention instead to \nl{factory}, allowing it successfully predict the coreference link with \nl{the four-story building}.

The task-specific nature of the attention mechanism is also illustrated in Example 4. The model generally pays attention to coordinators more than the content of the coordination, since coordinators, such as \nl{and}, provide strong cues for plurality.

The model is capable of detecting relatively long and complex noun phrases, such as \nl{a region of central Italy bordering the Adriatic Sea} in Example 2. It also appropriately pays attention to \nl{region}, showing that the attention mechanism provides more than content-word classification. The context encoding provided by the bidirectional LSTMs is critical to making informative head word decisions.

\paragraph{Weaknesses}
A benefit of using neural models for coreference resolution is their ability to use word embeddings to capture similarity between words, a property that many traditional feature-based models lack. While this can dramatically increase recall, as demonstrated in Example 1, it is also prone to predicting false positive links when the model conflates paraphrasing with relatedness or similarity. In Example 3, the model mistakenly predicts a link between \nl{The flight attendants} and \nl{The pilots'}. The predicted head words \nl{attendants} and \nl{pilots} likely have nearby word embeddings, which is a signal used---and often overused---by the model. The same type of error is made in Example 4, where the model predicts a coreference link between \nl{Prince Charles and his new wife Camilla} and \nl{Charles and Diana}, two non-coreferent mentions that are similar in many ways. These mistakes suggest substantial room for improvement with word or span representations that can cleanly distinguish between equivalence, entailment, and alternation.

Unsurprisingly, our model does little in the uphill battle of making coreference decisions requiring world knowledge. In Example 5, the model incorrectly decides that \nl{them} (in the context of \nl{let the rescuer locate them}) is coreferent with \nl{some ships}, likely due to plurality cues. However, an ideal model that uses common-sense reasoning would instead correctly infer that a rescuer is more likely to look for \nl{the man overboard} rather than the ship from which he fell. This type of reasoning would require either (1) models that integrate external sources of knowledge with more complex inference or (2) a vastly larger corpus of training data to overcome the sparsity of these patterns.

\section{Conclusion}
We presented a state-of-the-art coreference resolution model that is trained end-to-end for the first time. Our final model ensemble improves performance on the OntoNotes benchmark by over 3 F1 without external preprocessing tools used by previous systems. We showed that our model implicitly learns to generate useful mention candidates from the space of all possible spans. A novel head-finding attention mechanism also learns a task-specific preference for head words, which we empirically showed correlate strongly with traditional head-word definitions. 

While our model substantially pushes the state-of-the-art performance, the improvements are potentially complementary to a large body of work on various strategies to improve coreference resolution, including entity-level inference and incorporating world knowledge, which are important avenues for future work.

\subsection*{Acknowledgements}
The research was supported in part by DARPA under the DEFT program (FA8750-13-2-0019), the ARO (W911NF-16-1-0121), the NSF (IIS-1252835, IIS-1562364), gifts from Google and Tencent, and an Allen Distinguished Investigator Award. We also thank the UW NLP group for helpful conversations and comments on the work.

\bibliography{main}
\bibliographystyle{emnlp_natbib}

\end{document}